\documentclass{article}

\usepackage[final,nonatbib]{neurips_2023}
\usepackage{graphicx}

\usepackage[utf8]{inputenc} 
\usepackage[T1]{fontenc}    
\usepackage{hyperref}       
\usepackage{url}            
\usepackage{booktabs}       
\usepackage{amsfonts}       
\usepackage{nicefrac}       
\usepackage{microtype}      
\usepackage{xcolor}         
\usepackage{verbatim}
\usepackage{enumitem}

\title{DeepThought: An Architecture for Autonomous Self-motivated Systems}

    \author{%
  Arlindo L. Oliveira\\
  INESC-ID \&  ELLIS Lisbon Unit \\
  Instituto Superior Técnico, ULisboa, Portugal\\
  \texttt{arlindo.oliveira@tecnico.ulisboa.pt} \\
  \And
  Tiago Domingos\\
  MARETEC \\
  Instituto Superior Técnico, ULisboa, Portugal\\
  \texttt{tdomingos@tecnico.ulisboa.pt} \\
  \And
  Mário Figueiredo\\
  Inst. Telecomunicações \&  ELLIS Lisbon Unit \\
  Instituto Superior Técnico, ULisboa, Portugal\\
  \texttt{mario.figueiredo@tecnico.ulisboa.pt} \\
  \And
  Pedro U. Lima\\
  ISR \&  ELLIS Lisbon Unit \\
  Instituto Superior Técnico, ULisboa, Portugal\\
  \texttt{pedro.lima@tecnico.ulisboa.pt} \\
}

\begin{document}

\maketitle

\begin{abstract}
The ability of large language models (LLMs) to engage in credible dialogues with humans, taking into account the training data and the context of the conversation, has raised discussions about their ability to exhibit intrinsic motivations, agency, or even some degree of consciousness. We argue that the internal architecture of LLMs and their finite and volatile state cannot support any of these properties. By combining insights from \textit{complementary learning systems},  \textit{global neuronal workspace}, and \textit{attention schema} theories, we propose to integrate LLMs and other deep learning systems into an architecture for cognitive language agents able to exhibit properties akin to agency, self-motivation,  even some features of meta-cognition.
\end{abstract}

\section{Introduction and motivation}

The success of LLMs, based on the transformer architecture \cite{vaswani2017attention} and trained on very large datasets \cite{brown2020language,chowdhery2022palm}, has raised significant questions about whether such systems could exhibit some characteristics of higher-level natural intelligence: agency, self-motivation, even some aspects of consciousness. In fact, when equipped with long contexts, LLMs can engage in dialogues where their utterances can easily be mistaken for true signs of agency or sentience. For the purposes of this paper, we consider an LLM to be an autoregressive language model that predicts the next token by computing its conditional probability given the tokens in the context. More formally, language models learn a distribution $P(w_i \mid w_{i-1},...,w_{i-c})$, where $c$ is the context length. The model can then be used by sampling from the learned distribution, one token at a time. The use of \textit{reinforcement learning from human feedback} (RLHF) or fine-tuning techniques does not change the fundamental properties of language models, although it can significantly improve their performance. Given the sometimes surprising emergent properties of LLMs \cite{wei2022emergent}, there has been significant disagreement on whether scaling up these models may lead to the emergence of agency and purpose. In this position paper, we argue that simply scaling up LLMs is unlikely to lead to the emergence of the hallmarks properties of true intelligence, namely self-motivation and agency. However, integrating LLMs with other neural network-based architectures for vision and interaction with the external world as modules of a larger system may yield much richer behaviors that, if observed in humans or animals, are described as attention, agency, self-motivation, and even consciousness. 

Agency refers to the capacity of an entity to act intentionally, make choices, and exert control over its actions and decisions \cite{SEP}. An agent has the power to make meaningful choices that influence its state and that of the world around it. 
Agency is therefore necessarily associated with entities with internal states (memory) and capable of making intentional, self-motivated choices based on those states. 

LLMs have only a limited state, corresponding to the sequence of tokens in the context. The internal weights and other parameters do not store state information, as they are fixed after training and during interaction with the model. Since the context is finite, it provides only a limited state. Furthermore, every time an LLM starts a new conversation, it has an empty context, thereby ignoring all previous conversations.  It is thus clear that LLMs, even if scaled up to include much more training data, cannot, by themselves, exhibit the features that characterize agency and intrinsically motivated systems.

Agency implies the ability of a system (the agent) to act purposefully and make choices to achieve goals. It thus implies a degree of autonomy and the ability to decide based on some form of cognition and processing. While the criteria for agency can vary depending on the context and perspective, there are several characteristics often associated with agents, including:
autonomy (1), sensing (2), information processing (3), goal-oriented behavior (4), decision-making (5), adaptability (6), ability to act (7), persistence (8), learning (9), and intent or purpose (10). LLMs, by themselves, being essentially stateless systems driven by prompts, clearly lack several of these properties, although they possess some to a degree. Most notably, they lack properties 1, 4, 8, 9, and 10, even though their ability to dialogue may simulate, in some cases, some of them. The fact that their weights are fixed after training means they are not able to learn from interactions, thereby lacking property 9 (although they exhibit some form of in-context learning, it vanishes when each conversation ends). However, they are able to perform some limited sensing of the environment (by reading the input tokens), adapt to the input received so far (by considering the tokens in the context), make decisions based on the inputs, and perform a limited set of actions (by generating output tokens). To endow systems with the missing abilities from this list, it is necessary to enrich these models with memory, the ability to adapt to new situations, and mechanisms to pursue goals.


\section{Language and Cognitive Language Agents}

The ever-increasing popularity and capabilities of LLMs \cite{brown2020language,chowdhery2022palm} led rapidly to the idea that they could be the central components of architectures for autotelic agents. Well-known cognitive architectures, such as SOAR \cite{laird2019soar}, which have been proposed and tested for several decades, can gain significant flexibility and power by incorporating language models as components. In fact, the use of language models addresses two key limitations of cognitive architectures: they are limited to domains that can be formally specified; they require many hand-coded rules \cite{sumers2023cognitive}. LLMs are a natural choice to address these two limitations since they operate in arbitrary settings and are extremely flexible.

In the most common usage to date, language models take text as input and generate text as output. The only state they maintain corresponds to the context, the limited history of the interaction with the user, possibly augmented by an initial prompt that guides the behavior. A more sophisticated use of LLMs places them in a feedback loop with the outside world by converting observations into text and using the LLMs' output to choose actions, thus creating language agents \cite{sumers2023cognitive}. More advanced architectures, known as cognitive language agents, add additional components such as goal prioritization mechanisms (e.g., AutoGPT \cite{ject1297961}), code generation and storage \cite{wang2023voyager}, memory, and other symbolic reasoning abilities \cite{colas2023augmenting}. For instance, generative agents \cite{park2023generative} use an architecture that includes a memory stream and a retrieval mechanism that, when coupled with the abilities of GPT-4, leads to agency-revealing behavior that resembles that of human beings. 

Several of these proposals have shown that cognitive language agents exhibit relevant properties that are characteristic of agency, including autonomy, sensing, goal-oriented and autotelic behavior, and adaptability. However, we believe they fall short of achieving the full potential of self-motivated agents because the goals they pursue are either imposed externally or the result of random exploration of the state space and they lack attention mechanisms and meta-cognition (the ability to introspect their own internal cognition mechanisms).

We argue that meta-cognition and autotelic behavior require the use of \textit{attention}, \textit{i.e.},  the ability to focus the available resources on specific parts of the external input or of the internal state. This possibility to focus on (attend to) specific sub-parts of the available data is also behind our ability to break long-term goals into self-generated short-term goals and to adopt the behavior that leads to the fulfillment of the resulting sub-goals. Our proposal is inspired by a number of theories of attention and conscious behavior.

\section{Theories of attention and conscious behavior}

\textit{Complementary learning systems} theory (CLS) \cite{marr1971simple,mcclelland1995there} is a cognitive neuroscience theory that proposes a framework for understanding how the human brain processes and stores information, particularly in the context of memory and learning. This theory suggests that the brain employs two distinct and complementary systems for memory and learning: the hippocampus and the neocortex.

Rapid learning takes place in the hippocampus, in particular in the \textit{medium temporal lobe} (MTL), which is thought to be responsible for fast, episodic, and one-shot learning. It quickly encodes new information and experiences and plays a key role in declarative memory. This system also helps in differentiating and encoding similar but distinct experiences, preventing interference between them. The hippocampus adapts rapidly to new learning situations and is crucial for the formation of new episodic memories.
Slow learning happens in the neocortex, which is associated with slower, more gradual learning. This system consolidates information over time and is responsible for the storage and retrieval of semantic or factual knowledge, and well-established memories. It is also involved in extracting commonalities and general principles from individual experiences.

\begin{figure}[hb]
\begin{minipage}{.5\textwidth}
  \centering
  \includegraphics[width=6cm]{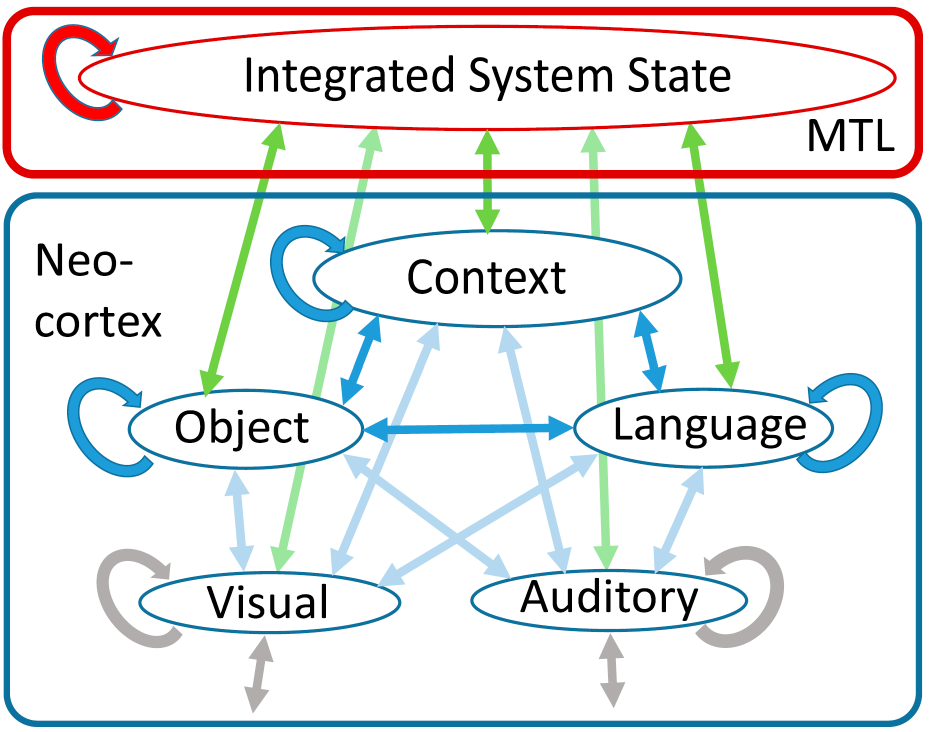}
\end{minipage}
\begin{minipage}{.5\textwidth}
  \centering
  \includegraphics[width=6cm]{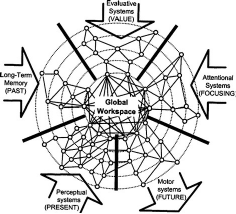}

\end{minipage}
  \caption{Left: The brain understanding system (reproduced from \cite{mcclelland2020placing}). The red arrow represents fast-learning connections while the green arrows represent slow-learning connections. Right: an illustration of the global workspace theory (reproduced from \cite{dehaene1998neuronal}). When information is brought into the global workspace, it becomes conscious and is broadcast to all the (unconscious) processors.   \label{BrainSystem}}
\end{figure}

CLS theory argues that these two systems work together in a complementary manner to support memory and learning. Initially, when we encounter new information, the hippocampus rapidly encodes it, but these memories are fragile and degrade over time. The neocortex gradually consolidates these memories, making them more stable and resistant to interference. A key idea behind the CLS theory is that the rapid learning facilitated by the hippocampus is essential for adapting to novel situations, while the slower, more stable neocortical learning is responsible for building a reliable knowledge base. This duality allows humans to acquire new knowledge quickly and integrate it into their existing understanding of the world, something sorely lacking in current deep learning systems. Recently, McClelland et al. \cite{mcclelland2020placing} suggested an architecture for the \textit{brain understanding system}, depicted in Fig. \ref{BrainSystem} (left). While this figure aims to represent the fundamental architecture of biological brains, it can also serve as a blueprint for autotelic artificially intelligent systems.

\textit{Global workspace theory} (GWT) (which is closely related to the \textit{global neuronal workspace} -- GNW -- theory) \cite{dehaene1998neuronal,baars2007global,dehaene2014consciousness} addresses distinct aspects of cognitive processing, by attempting to explain how the brain processes information and achieves conscious awareness.
While CLS theory primarily deals with the organization of memory systems and how different brain regions work together to support memory and learning, GWT/GNW is a theory of consciousness and attention, primarily concerned with how information is selected and broadcast to conscious awareness. GWT/GNW postulates a "global workspace" in the brain where information from various cognitive processes (perception, memory retrieval, decision-making, etc.) competes for attention and conscious processing.
In GWT/GNW, information from different sources is integrated and made available to conscious awareness (Figure \ref{BrainSystem}, right). GNW postulates that, apart from specialized cortical areas that process different types of information, there is a distributed network of GNW neurons with long-range connections, which selectively influence specific processing neurons and act as a communication hub, receiving input from lower levels and transmitting guidance to different processing regions.

Although they have different assumptions and postulate different mechanisms, CLS and GWT/GNW are clearly related. This relationship suggests that combining their ideas can provide a more comprehensive account of cognitive processing, especially in the context of memory and conscious awareness. In particular, GWT/GNW may help explain how the information processed and consolidated by the CLS's neocortical system becomes accessible to conscious awareness. When a memory or piece of knowledge is retrieved and brought into the global workspace, it can be consciously experienced. GWT/GNW's notion of attention and global workspace can also be related to CLS in terms of how attention processes affect learning and memory encoding. Information that receives attention is more likely to be rapidly encoded by the hippocampus, initiating or strengthening the learning process.

The role of the MTL in CLS theory could, therefore, be somehow related to the role of the global workspace in GWT/GNW. The fact that the mechanisms proposed by the two theories do not fully coincide may be due to our limited knowledge of the process but does not preclude the existence of a deep relationship between the two postulated mechanisms. In fact, the MTL may well play the role of a sensory hub where visual features are ``bound” into single, conscious (reportable) gestalts and widely distributed to the neocortex. Recent work has shown that concept cells, which are found in the medial temporal lobe of humans and activate when subjects perceive or think about specific concepts, fire when concepts are held in working memory \cite{mashour2020conscious}.

Another theory that is relevant in the context of this discussion is Graziano's \textit{attention schema theory} (AST) \cite{graziano2015attention}. This theory posits that the brain constructs a simplified internal model, or "schema," of attention, which enables the brain to attribute awareness to itself, creating a self-referential understanding of attention. Graziano argues that consciousness arises from the brain's ability to model its own attentional processes. Although they are independent, it is possible to establish a connection between AST and GWT/GNW \cite{graziano2015attention}. In the context of the attention schema theory, the global workspace serves as a platform for the brain to represent and manipulate the internal model of attention, contributing to the overall experience of consciousness.

\section{A biologically inspired architecture for autotelic systems}

Driven by evidence from biological systems and by the known behavior and limitations of existing models, we propose an architecture for intrinsically motivated systems that do open-ended learning. Our proposal is founded on three essential concepts. The first dictates the necessity of a two-system architecture, informed by CLS and GWT/GNW theories, as well as dual process theories \cite{frankish2010dual,kahneman2011thinking}, which posit that there are two distinct information processing modes in the brain: one is fast, unconscious and high-bandwidth; the other is slow, deliberate and conscious.  The second concept is that the behavior of the system is controlled by an attention schema \cite{graziano2015attention}, which takes into account available data to direct the focus of attention of the supervisor module. The third concept posits that the driving force behind a system capable of continuous learning is its inherent ``desire" to enhance its understanding/model \cite{schmidhuberdriven} and control \cite{barto2013intrinsic} of the world, thereby sustaining goals that promote ongoing improvement in both the short and long term. This proposal combines ideas as diverse as 
\begin{itemize}[leftmargin=0.3cm]
\item \textit{libido sciendi} ("passion for knowledge") \cite{fages2018introduction}, which provides the driver for learning;
\item Friston's free energy principle for the brain \cite{friston2010free} (which posits that the brain aligns its internal model of the world with the perceived external world by making predictions based on internal models of actions and updating them using sensory inputs), which provides the direction for learning; 
\item Popper's epistemological theory of falsificationism \cite{popper2014conjectures}, whereby actions (experiments) are performed on the outside world to test "risky" hypotheses, \textit{i.e.}, statements about the world of high generality, maximizing the rate of learning.
\end{itemize}

\begin{figure}[hbt]
  \centering
  \includegraphics[width=12.75cm]{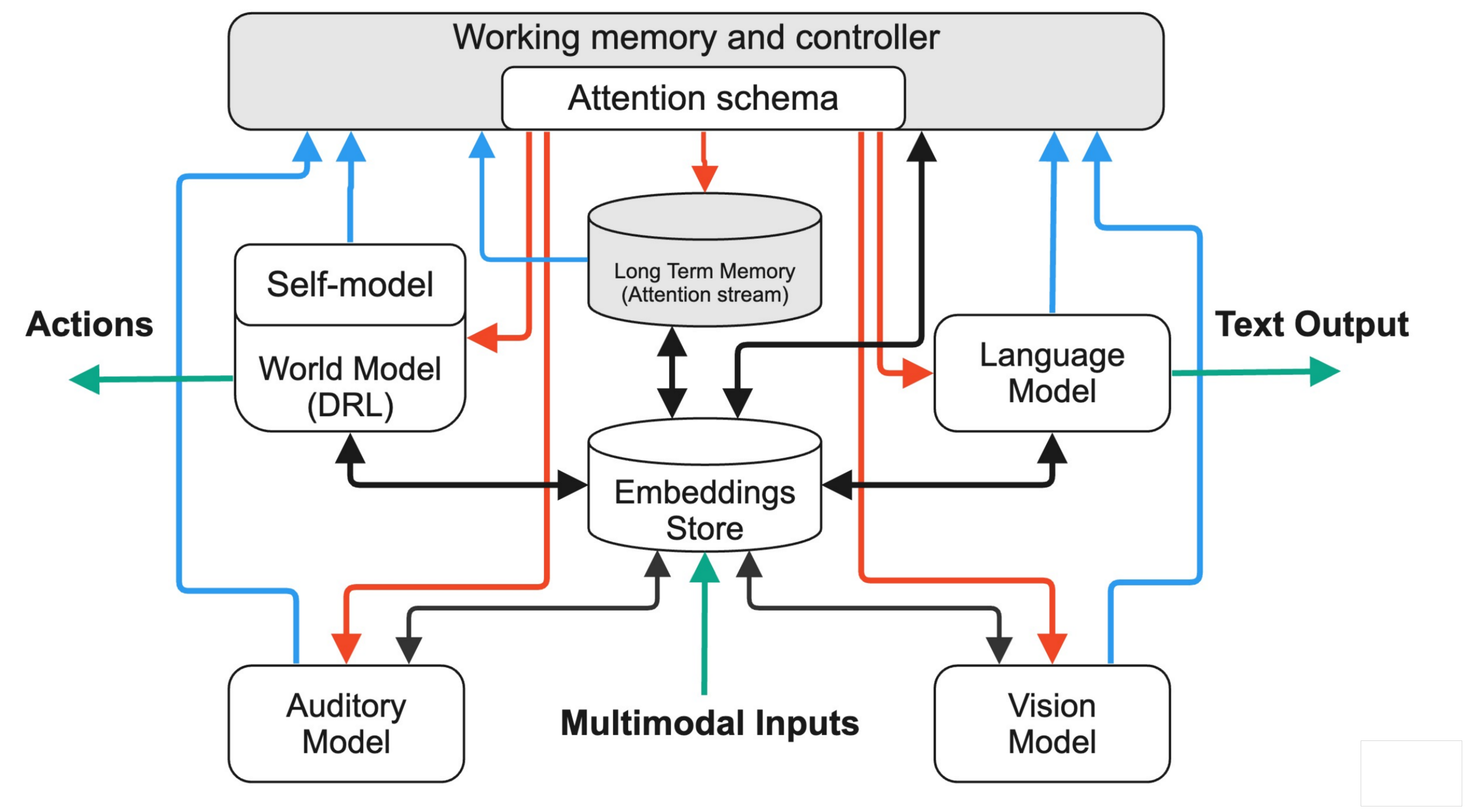}
  \caption{The DeepThought architecture: green arrows represent inputs and outputs, blue arrows represent module outputs, red arrows represent attention and control signals, and black arrows represent bidirectional interactions between modules and the embeddings store. The working memory and controller, which includes the attention schema, define the next inputs to the different modules.}
  \label{DeepThought}
\end{figure}

Figure \ref{DeepThought} depicts the proposed \textit{DeepThought} architecture, which includes several specific modules that play relevant roles. The \textbf{supervisor module}, at the top, enables short-term attention and memory and acts as the medium temporal lobe in the CLS theory, displaying fast adaptability, and controlling the global attention mechanism posited by GWT/GNW and AST. It receives inputs from the (system 1) language, vision, auditory, and world models at the bottom and focuses attention on specific inputs or on specific outputs of these models. As in GWT, tokens that deserve attention are broadcast to all the modules for explicit processing. The \textbf{deep reinforcement learning} (DRL) module uses the inputs from the external world, under the control of the \textbf{attention schema}, and compares them with the internal world and self models, adjusting, as required, the model parameters. 

As proposed by other authors \cite{schmidhuber2006developmental}, the DRL module is rewarded for action sequences that improve the predictions of the world model. These actions are suggested to the supervisor module, which may or may not decide to focus attention on them. The \textbf{language}, \textbf{auditory}, and \textbf{vision} modules, under the control of the attention schema, process information received from the outside world and/or from the supervisor (via the \textbf{embeddings store}). The embeddings in this store are derived by the joint encoding of external and internal multimodal token streams, as is done in  CLIP \cite{radford2021learning}. 

The architecture also includes several memory components: a short-term \textbf{working memory}, which is part of the supervisory module, and a long-term episodic memory, which corresponds to the attention stream and is used to store explicit knowledge about the past.  The succession of tokens that are the focus of attention is stored in \textbf{long-term memory} (the attention stream) for future retrieval and also to make meta-cognition possible. The system can describe the flow of reasoning, listing the series of sensations and actions that were the focus of attention. Semantic and procedural memory, on the other hand, are stored essentially in the language, vision, auditory, and world models, and can only be changed slowly, by adapting their parameters. We do not detail here the exact training mechanism that can be used to adjust the model parameters, mimicking the slow learning processes that take place in the neocortex, according to the CLS theory.

\vspace{-0.5cm}

\section{Conclusions}

Inspired by the current state of the art in cognitive language agents and by the GWT/GNW, AST, and CLS theories of cognition, we have proposed \textit{DeepThought}, a two-system architecture for cognitive language agents that exhibit important characteristics of agency. The architecture includes several novel components, such as the integration of an attention schema, a clear definition of the role of the supervisory module, and the way system 1 and system 2 processes are combined. 

\section*{Acknowledgements}
We acknowledge financial support from the Recovery
and Resilience Fund towards the Center for Responsible AI
project (Ref. C628696807-00454142), the Foundation for Science
and Technology (FCT) through Project PRELUNA - PTDC/CCIINF/
4703/2021 and the FCT multiannual financing of INESC-ID (Ref.
UIDB/50021/2020) and for LARSyS (Ref UIDB/50009/2020). We also thank the anonymous reviewers for their insightful and constructive suggestions.

\bibliographystyle{unsrt}
\bibliography{references}


\end{document}